\documentclass[runningheads]{llncs}

\usepackage{eccv}

\usepackage{eccvabbrv}

\usepackage{graphicx}
\usepackage{booktabs}

\usepackage[accsupp]{axessibility}  

\usepackage{hyperref}

\usepackage{orcidlink}

\usepackage{booktabs}
\usepackage{epsfig}
\usepackage{physics}
\usepackage{caption}
\usepackage{subcaption}
\usepackage{multirow}
\usepackage{array}
\usepackage{enumerate}
\usepackage{algorithm}
\usepackage{algorithmic}
\usepackage{makecell}
\usepackage{wrapfig}

\usepackage{xspace}
\usepackage{amssymb}
\usepackage{siunitx}

\AtBeginDocument{\RenewCommandCopy\qty\SI}

\makeatletter
\DeclareRobustCommand\onedot{\futurelet\@let@token\@onedot}
\def\@onedot{\ifx\@let@token.\else.\null\fi\xspace}

\def\eg{\emph{e.g}\onedot} 

\def\ie{\emph{i.e}\onedot}

\AtEndPreamble{
    \usepackage[capitalize]{cleveref}
    \crefname{section}{Sec.}{Secs.}
    \crefname{table}{Tab.}{Tabs.}
    \crefname{chapter}{Chapter}{Chapters}
    \crefname{figure}{Fig.}{Figs.}
    \crefname{algorithm}{Alg.}{Algs.}
}

\newcommand{\datasetname}{RSP3D\xspace}

\begin{document}

\title{Prosthesis-Aware 3D Human Pose Estimation:
A Dataset and Benchmark for RSP Users}

\titlerunning{RSP3D}

\author{Yilin Wen\inst{1}\orcidlink{0000-0002-5981-1276} \and
Kechuan Dong\inst{1} \and
Fumiya Suginaka\inst{1}\and
\\
Ken Endo\inst{2}\orcidlink{0009-0006-3574-8563}\and
Yusuke Sugano\inst{1}$^*$\orcidlink{0000-0003-4206-710X}}

\authorrunning{Y.~Wen et al.}

\institute{The University of Tokyo, Tokyo, Japan  \\
\and
Sony Computer Science Laboratories, Tokyo, Japan\\
\email{\{fylwen,kchdong,fumisugi,sugano\}@iis.u-tokyo.ac.jp}\\
\email{kene@csl.sony.co.jp}}

\maketitle
\let\thefootnote\relax\footnotetext{$^*$Corresponding author}
\begin{abstract}
Recovering 3D human body motion from video is important for applications such as rehabilitation assessment and sports performance evaluation. 
For prosthesis users, this requires capturing both natural body joints and the geometry of the prosthetic device, a challenge that existing methods are not designed to address. 
Model-based estimators rely on body models trained on non-amputee individuals and cannot represent prosthesis geometry, while model-free methods lack body kinematic priors and are unreliable under occlusion. 
This challenge is particularly prominent for users of running-specific prostheses (RSPs), where the RSP has a complex curved geometry and moves dynamically during exercise. 
To fill this gap, we collect \datasetname, the first 3D dataset of RSP users, covering essential daily-life and exercise actions from participants with varied amputation conditions, using a multi-camera marker-based motion capture setup. 
We formally define the task of prosthesis-aware 3D pose estimation, evaluate representative methods in a zero-shot setting, and confirm their individual limitations. 
We further propose a hybrid baseline combining model-based body joint estimation with model-free RSP shape recovery, establishing a starting point for future research.
Our project page is available at \url{https://ut-vision.github.io/RSP3D/}

\keywords{3D Human Pose Estimation \and Prosthesis Reconstruction \and Dataset and Benchmark \and Running-Specific Prosthesis}

\end{abstract}

\section{Introduction}

Accurately recovering 3D human body motion from videos is a key technology for understanding human activities.
For prosthesis users, capturing the 3D positions of both natural body joints and prosthetic components is essential for applications such as rehabilitation assessment and sports performance evaluation.
Such prosthesis-aware 3D body recovery can directly improve the quality of life for prosthesis users and contribute to building a more inclusive society. 
While professional athletes typically have access to coaches who can monitor their movement and provide feedback, for the majority of running-specific prostheses (RSP) users who train independently or in recreational settings, such expert guidance is not always available. 
Automated prosthesis-aware 3D analysis can therefore fill this gap, providing objective motion feedback that supports training and rehabilitation without requiring direct expert supervision.
However, existing methods have not been designed to handle prosthesis users, and there is a clear lack of both data and methods for this purpose.

\begin{figure}[!t]
\centering
\includegraphics[width=.99\textwidth]{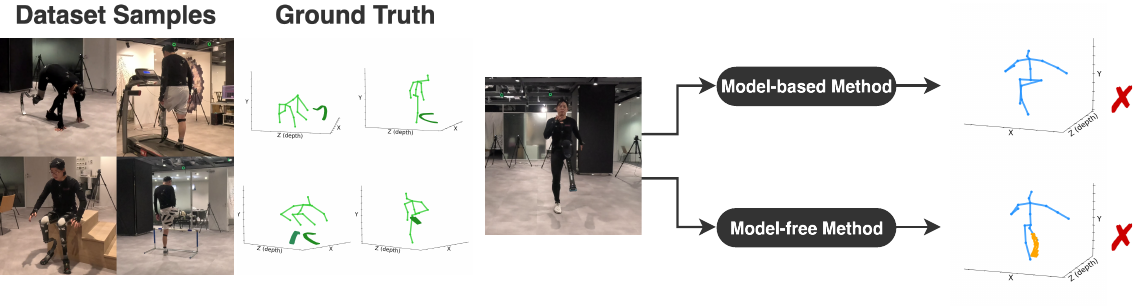}
\caption{Samples of collected data (\textit{left}) and challenges faced by existing works in handling pose estimation for running-specific prostheses (RSP) users (\textit{right}).}\label{fig:fig1}
\end{figure}

This task poses a unique challenge: part of the human body is replaced by an object that does not follow human anatomical structure, and prostheses lack shared statistical geometric structure across different users.
This breaks the fundamental assumption shared by existing approaches, and neither model-based nor model-free methods can solve it alone (see~\cref{fig:fig1}).
Model-based 3D human pose estimators~\cite{zhu2023motionbert,yang2026sam3dbody} rely on predefined anatomical models~\cite{ionescu2013human3,SMPL:2015,MHR:2025} trained on large-scale datasets of non-amputee individuals, and by design cannot generalize to bodies outside this learned distribution. 
This means that while they can robustly estimate body joint positions even under occlusion, they cannot represent prosthesis geometry.
Moreover, prostheses vary widely in shape, length, and attachment depending on the individual's amputation condition and activity, making it difficult to define a common geometric prior.
Conversely, model-free approaches~\cite{xiao2024spatialtracker,xiao2025spatialtrackerv2,chen2026sam} can reconstruct arbitrary 3D shapes without relying on predefined templates, making them capable of recovering prosthesis geometry.
However, without body-specific priors, they often fail under occlusion and cannot capture the correct kinematics of body joints.
In other words, this task inherently requires integrating two distinct approaches: one that understands the human body and one that can handle arbitrary shapes.

In this study, we address this challenge by constructing \datasetname, the first dataset that captures 3D poses of prosthesis users, with a particular focus on individuals using running-specific prostheses (RSPs).
Our dataset comprises essential actions for daily activities and specific drills for physical exercise, collected from diverse participants spanning a range of ages, amputation conditions, and RSP experience.
We use an indoor environment equipped with 16 GoPro RGB cameras and a marker-based motion capture system, providing 3D annotations of both natural body joints and RSP shapes.
With this dataset, we formally define the task of prosthesis-aware 3D human pose estimation, which involves recovering both natural body joints and keypoints representing the prosthesis shape in 3D space from video observations.

We then use the collected data for zero-shot evaluation, considering representative model-based and model-free methods.
For model-based approaches, we evaluate SAM3D Body~\cite{yang2026sam3dbody}, MotionBERT~\cite{zhu2023motionbert}, and the only existing amputation-aware method AJAHR~\cite{cho2025ajahr}.
For model-free approaches, we evaluate SpatialTrackerV2~\cite{xiao2025spatialtrackerv2}, which enables 3D point tracking and scene reconstruction from video.
Our evaluation confirms their individual limitations discussed above.
Based on these findings, we propose a hybrid baseline that combines the two approaches: we use model-based methods for robust body joint estimation and model-free methods for flexible prosthesis shape recovery, aligning their outputs through scale and position adjustment, therefore providing a reference for future research in prosthesis-aware 3D pose estimation.

Our contributions are summarized as follows: 
\begin{itemize}
\itemsep0em
\item We introduce the task of prosthesis-aware 3D human pose estimation, which requires recovering both natural body joints and prosthesis geometry from video observations.
\item We construct \datasetname, the first 3D dataset of RSP users, covering actions from daily activities to physical exercise, with 3D annotations for both body joints and RSP shapes, serving as a resource for zero-shot benchmarking.
\item We evaluate existing model-based and model-free methods, revealing their individual limitations, and propose a hybrid baseline that integrates both approaches to establish a starting point for this new research direction.
\end{itemize}

\section{Related Work}

\subsubsection*{3D Human Pose Estimation}

2D human pose estimation aims to detect body joint locations in image space. 
Mainstream solutions~\cite{fang2022alphapose,newell2016stacked,sun2019deep,xu2022vitpose} typically involve detecting the human bounding box and regressing the heatmap of a fixed set of body joints.
However, these methods face a specific challenge for prosthesis users, as joint detectors trained on non-amputee individuals may fail to localize joints near prosthetic limbs.
2D-to-3D lifting approaches convert 2D joint detections into 3D space~\cite{zhu2023motionbert,zheng20213d,martinez2017simple}.
For prosthesis users, these methods inherit errors from 2D detection, and the lifting models themselves assume a complete set of body joints, making them sensitive to missing or incorrectly detected joints.

Direct 3D estimation methods recover body joints and mesh directly from images~\cite{kolotouros2019learning,kanazawa2018end,goel2023humans,yang2026sam3dbody,kocabas2020vibe,li2025genmo,shin2024wham}.
By training on large datasets that cover a variety of body poses and environments, these solutions learn consistent spatiotemporal relationships among different joints, enabling robust performance in diverse settings and accurate predictions even when joints are occluded.
Recent works~\cite{cseke2025pico,xie2026cari4d,wen2025reconstructing,xie2024template} have further extended this to reconstructing humans alongside the objects they interact with.
However, these models cannot represent bodies outside their training distribution, making them unsuitable for prosthesis users.

Across these approaches, the fundamental limitation lies in the assumption of a complete set of body joints.
While 2D detectors struggle to localize joints near prosthetic limbs, lifting-based and direct 3D methods are further constrained by body models trained exclusively on non-amputee individuals.
As a result, none of these methods can capture amputations or the geometry of RSPs.

\subsubsection*{Model-free 3D Reconstruction}

Model-free solutions recover 3D structures of arbitrary objects from image observations or 2D keypoints, without relying on predefined shape templates.
Earlier research tackled this using non-rigid structure-from-motion techniques~\cite{bregler2000recovering,dai2014simple,kumar2020non}, which recover 3D structure from 2D point trajectories of deforming objects.
3D-LFM~\cite{dabhi20243d} proposes a unified framework for freeform 2D-to-3D lifting that captures object deformation while preserving point permutation equivariance, showing generalizability to novel objects. 
However, these methods rely heavily on the accurate 2D detection of keypoints, and often struggle with resolving depth ambiguity due to the sparse nature of inputs.

More recently, foundational backbones for scene depth and 3D pointwise features~\cite{yang2024depth,oquab2024dinov2,wang2025vggt,wang2024dust3r} have enabled unified solutions for dynamic 3D/4D point correspondence, supporting simultaneous scene reconstruction and point tracking~\cite{zhang2025monst3r,chen2025easi3r,wang2025c4d,feng2025st4rtrack,zhang2026efficiently,xiao2024spatialtracker,xiao2025spatialtrackerv2,wang2025shape,zhang2025tapip3d}. 
Without relying on predefined object templates, these methods can recover 3D shapes corresponding to the 2D detection and segmentation of body parts and RSPs, thus providing a feasible solution for our task. 
However, this flexibility complicates the incorporation of body kinematic priors, reducing the accuracy of natural body joint estimation, as these methods cannot exploit the kinematic relationships among joints.

\subsubsection*{Pose Estimation for Prosthesis Users}

\begin{table}[!tp]
\caption{Comparison with existing datasets.}\label{tab:dataset}
\centering
\resizebox{.99\linewidth}{!}{
\begin{tabular}{c|cccccc}
\hline
\textbf{Dataset} & \textbf{Data Source} & \makecell{\textbf{\#}\\\textbf{Participants}} & \makecell{\textbf{\#}\\\textbf{Frames}} & \makecell{\textbf{Prosthesis}\\\textbf{Users}} & \makecell{\textbf{Prosthesis}\\\textbf{Ann.}} & \makecell{\textbf{Body}\\\textbf{Ann.}} \\
\hline
Human3.6M~\cite{ionescu2013human3} & Real-world Collection & 11 & 3.6M & $\times$ & - & 3D \\
MPI-INF-3DHP~\cite{mehta2017monocular} & Real-world Collection & 8 & 1.3M & $\times$ & - & 3D \\
3DPW~\cite{von2018recovering} & Real-world Collection & 5 & 51k & $\times$ & - & 3D \\
LDPose~\cite{ying2025ldpose} & Web Images/Videos & - & 28k & \checkmark & - & 2D \\
InclusiveVidPose~\cite{du2026inclusivevidpose} & Web Videos & - & 327k & \checkmark & - & 2D \\
ProGait~\cite{yin2025progait} & Real-world Collection & 4 & 412 clips & \checkmark & 2D & 2D \\
A3D~\cite{cho2025ajahr} & Synthetic Data & - & 1.0M & \checkmark & - & 3D \\
Ours & Real-world Collection & 6 & 5.6M & \checkmark & 3D & 3D \\
\hline
\end{tabular}}
\end{table}

Several recent works have addressed body pose estimation for prosthesis users.
LDPose~\cite{ying2025ldpose} and InclusiveVidPose~\cite{du2026inclusivevidpose} focus on limb deficiency-aware 2D pose estimation, where they collect data using web images and videos, and concentrate on detecting intact or residual body joints separately.
ProGait~\cite{yin2025progait} further considers the prosthesis shape in 2D space for gait analysis.
This study involves collecting data from human subjects with transfemoral prosthetic legs and examines 2D human pose estimation and object segmentation to facilitate gait classification.
AJAHR~\cite{cho2025ajahr} extends this to 3D by predicting amputation-aware body meshes. 
This is achieved by first classifying the body part amputation and generating a mesh reflecting this amputation status. 
Furthermore, they propose a synthesis pipeline to generate data for training and evaluation.

Despite these advances, existing solutions still cannot capture both body joints and prosthesis geometry in 3D space.
The lack of datasets tailored for this task remains a fundamental obstacle, limiting the development of dedicated methods (see \cref{tab:dataset}).
\section{Dataset}

We collect \datasetname, a motion dataset from prosthesis users focusing specifically on those using RSPs.
We capture and annotate both the 3D positions of natural body joints and the RSP shape, which has not been addressed in 3D in prior research.
This study was approved by the Institutional Review Board under approval number E2025ALS275.
Our data and code are accessible through our project page. 
Users must agree to the license and terms of use outlined on our project page and submit an access request form. 
Approved applicants will receive access to the videos and annotations.

\subsection{Recording Setup}
\begin{figure}[!t]
\centering
\includegraphics[width=.99\textwidth]{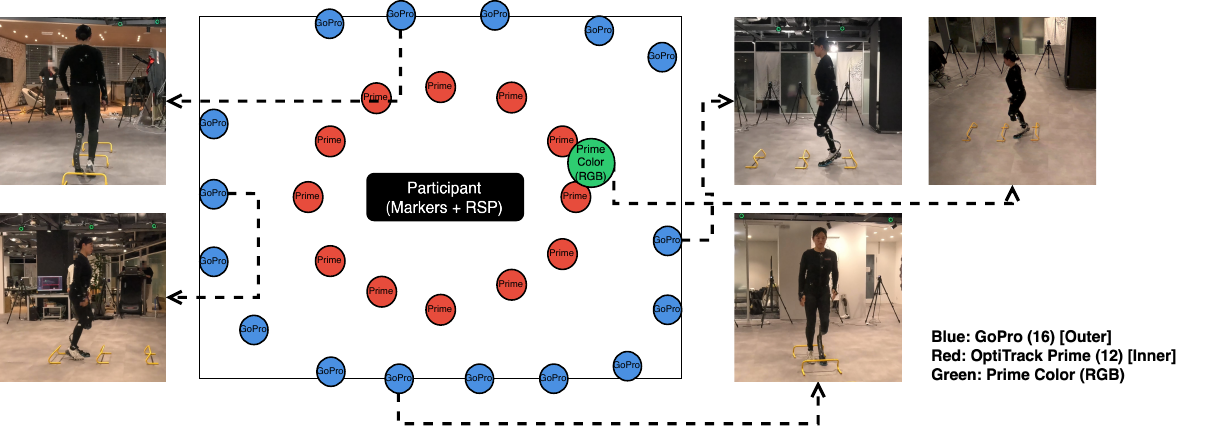}
\caption{Our recording setup.}\label{fig:recording}
\end{figure}

\paragraph{Camera Setup}
As shown in~\cref{fig:recording}, we conduct our dataset collection in an indoor laboratory, where we record using a combination of an OptiTrack motion capture system and 16 GoPro video cameras.
We place reflective markers on participants to capture their 3D positions.
The OptiTrack system employs 12 Prime cameras for marker tracking and one Prime Color camera for video recording.
The 16 GoPro cameras surround the recording area to capture video at 60 FPS from various angles.

\paragraph{Time Synchronization and Calibration}
We follow the official procedure to synchronize and calibrate the OptiTrack cameras, including hardware-based time synchronization and waving a calibration wand throughout the recording space.
To synchronize the OptiTrack system with the GoPro cameras, we employ a QR video method inspired by Ego-Exo4D~\cite{grauman2024ego}.
At the start of each recording session, we display a QR video to all GoPro cameras and the Prime Color camera, where each frame encodes a timestamp at a resolution of \SI{~0.033}{sec} (\ie, 30 FPS).
Each camera's time code is synchronized by detecting this QR video.
Through manual checks, we ensure synchronization accuracy within one frame at 60 FPS.
We calibrate the intrinsic parameters of each GoPro camera and the Prime Color camera using a ChArUco board~\cite{garrido2014automatic} via OpenCV, following the standard procedure of capturing the board from multiple viewpoints.
Extrinsic calibration among these cameras is conducted using another ChArUco board with a known physical size, establishing the transformation between each camera's coordinate system and the world coordinate system defined by the OptiTrack system.

\subsection{Action}
\begin{figure}[!t]
\centering
\includegraphics[width=.99\textwidth]{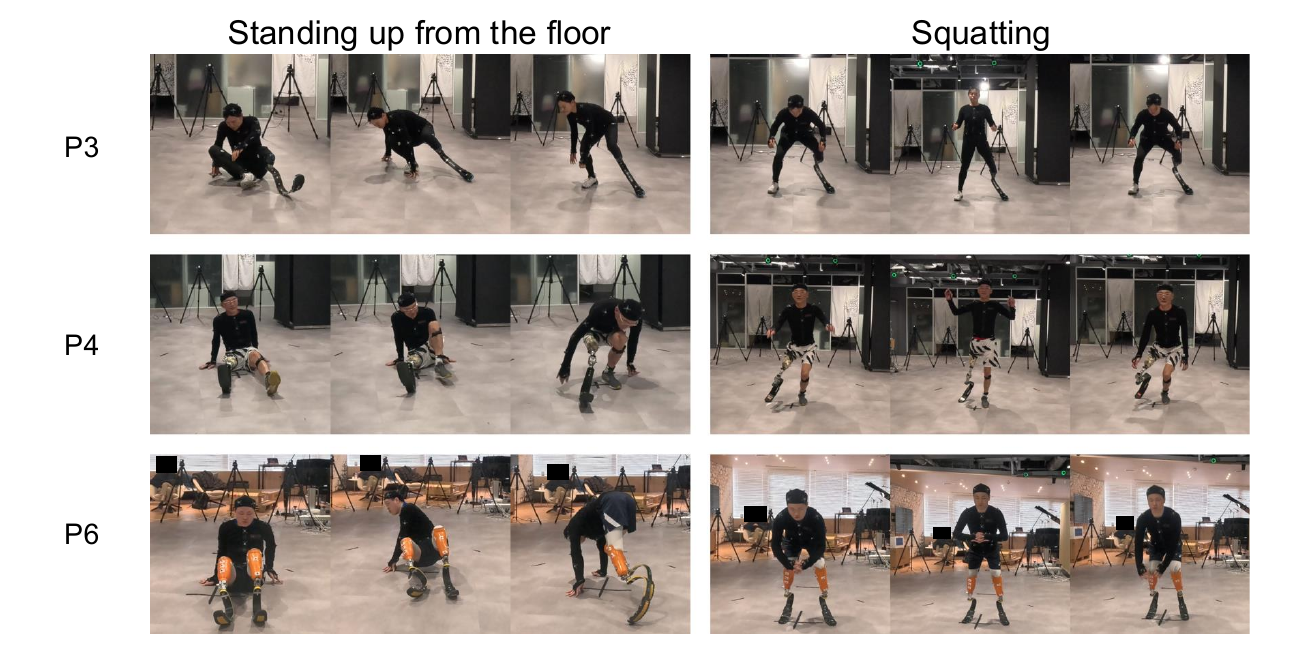}
\caption{Actions that highlight differences in motion patterns among participants with varying amputation conditions. 
We show more cases in the supplementary materials.}\label{fig:actions}
\end{figure}
We design the recording sessions in collaboration with two experienced prosthetists and orthotists to ensure that the selected actions are both practically relevant and representative of the challenges faced by RSP users.
To build an initial action list, we attended a training session for first-time RSP users led by the prosthetists, observing which movements were practiced and which proved difficult.
We then refined the list through a follow-up meeting with the prosthetists, incorporating their feedback on which actions were unexpectedly important for daily life and which presented particular difficulty for RSP users.

On one hand, in addition to basic activities like walking and jogging, we include actions crucial for daily life (\eg, \textit{body rotation in place}, \textit{standing up from the floor}), as well as those that test proficiency in using the RSP while maintaining balance (\eg, \textit{stepping over hurdles}, \textit{vertical jumps}).
On the other hand, we incorporate actions designed to highlight motion pattern differences among participants with varying amputation conditions (see \cref{fig:actions}).
For instance, noticeable differences are observed in actions like \textit{walking up stairs} and \textit{squatting} between individuals with below-knee and above-knee amputations, primarily due to the absence of the knee joint.
Similarly, users with left-foot versus right-foot amputations may demonstrate different strategies for \textit{standing up from the floor}, often shifting weight to the non-amputated side.
Furthermore, for users with bilateral amputations, movements such as \textit{hip rotation} introduce additional challenges in maintaining balance and body control.
A detailed description of these actions, along with examples, is provided in the supplementary materials.

\begin{figure}[!t]
\centering
\includegraphics[width=.99\textwidth]{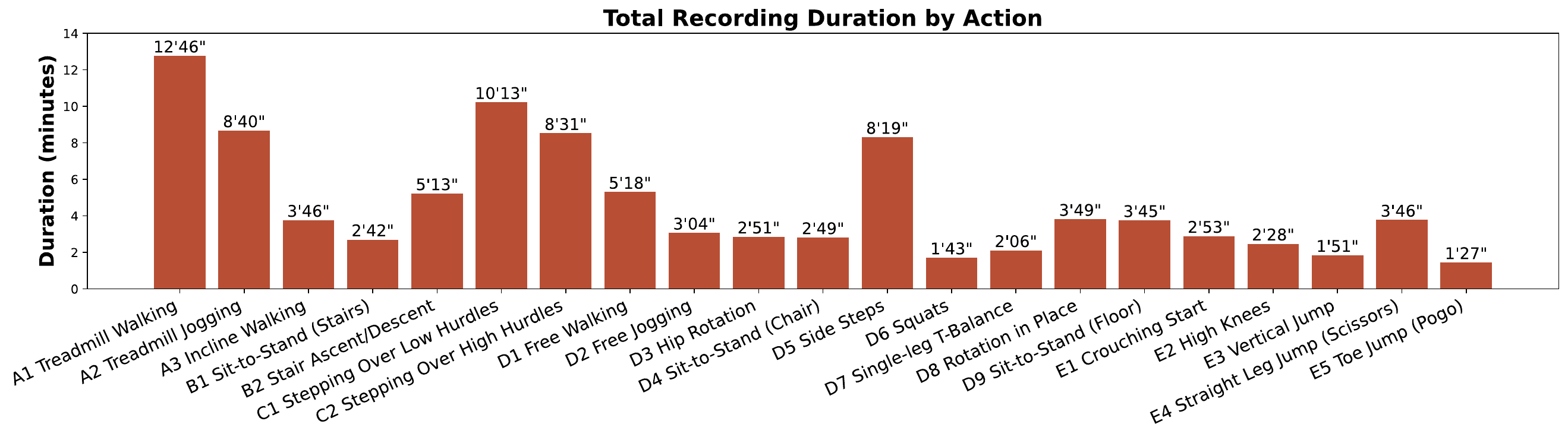}
\caption{Total recording duration by action.
We report on all 6 participants.}\label{fig:statistics}
\end{figure}

\subsection{Annotation}
As illustrated in~\cref{fig:recording}, we attach reflective markers to the body parts and RSPs of participants. 
OptiTrack cameras capture the positions of these markers within their world coordinate system, allowing us to reconstruct the 3D positions of natural body joints and the RSP surfaces.

\paragraph{Body Joints}
We use the 17-joint definition from the Human3.6M dataset~\cite{ionescu2013human3} to annotate the natural body joints.
For the upper body, including the hips, we use the upper body template provided by the OptiTrack system, following its guidelines for marker placement and joint position regression.
For the available knees and ankles, we place two markers symmetrically on the left and right sides of each available knee or ankle, using their midpoint as the corresponding annotation.

\paragraph{RSP Surface}
To capture the RSP geometry, we densely and symmetrically place markers on both sides of the RSP along its curve (\cref{fig:recording}).
Let the marker set be $\bar{\mathcal{M}}=\bar{\mathcal{M}}^1\bigcup\bar{\mathcal{M}}^2$, where $\bar{\mathcal{M}}^1$ and $\bar{\mathcal{M}}^2$ collect markers on each side of the RSP.
Each $\bar{\mathcal{M}}^i$ contains $n$ sequential markers $(\bar{\vb*{m}}^i_1,\ldots,\bar{\vb*{m}}^i_n)$ that are ordered from the near-body side to the near-floor side. 
As shown in \cref{fig:recording}, markers are placed at intervals of approximately \SI{5}{cm}, allowing for the capture of the RSP curvature.

We then construct the RSP surface $\bar{\mathcal{S}}$ using a ruled surface approach, which effectively captures the RSP curve shape and surface width.
Specifically, $\bar{\mathcal{S}}$ is defined by a family of lines $\bar{\vb*{r}}_u(v)$ that span between $\bar{\mathcal{M}}^1$ and $\bar{\mathcal{M}}^2$, connecting markers $\bar{\vb*{m}}^1_u, \bar{\vb*{m}}^2_u$ with the same index from each side.
$\bar{\mathcal{S}}$ is then expressed as
\begin{equation}
\bar{\mathcal{S}} = \bigcup_{u=1}^{n} \{\bar{\vb*{r}}_u(v) = (1-v)\bar{\vb*{m}}^1_u + v\bar{\vb*{m}}^2_u \mid v \in [0, 1]\}.
\end{equation}

\begin{table}[!t]
\centering
\caption{Information of participants.}\label{tab:participant}
\begin{tabular}{ccccc}
\midrule
ID & Age Group & Gender & Amputation Site & RSP Experience \\
\midrule
P1 & 30s & M & Below-knee (Right) & > 5 yrs \\
P2 & 40s & M & Above-knee (Left) & < 1 yr \\
P3 & 20s & F & Below-knee (Left) & > 5 yrs \\ 
P4 & 60s & M & Above-knee (Right) & > 5 yrs \\
P5 & 10s & F & Below-knee (Bilateral) & > 5 yrs \\
P6 & 10s & M & Below-knee (Bilateral) & < 1 yr \\
\midrule
\end{tabular}
\end{table}

\subsection{Statistics}
We collect data from 4 male and 2 female RSP users, ensuring diversity in age, amputation conditions, and years of RSP experience.
Since RSP users span a wide range of training backgrounds, from beginners who have recently adopted the device to experienced recreational runners, capturing this diversity is important for developing methods that generalize beyond elite users who have access to professional coaching.
We provide the demographic statistics in~\cref{tab:participant}, and a breakdown of recording lengths by action in~\cref{fig:statistics}.
Additionally, we compare \datasetname with existing related datasets in \cref{tab:dataset}.
Overall, our data collection offers a comprehensive resource for motion analysis of RSP users across different body conditions, including substantial recording lengths and a variety of camera viewpoints for diverse participants and actions.
\section{Methods and Evaluation Protocol}
\subsection{Task Definition}

As illustrated in \cref{fig:fig1}, our task input is a monocular video $\mathcal{V}=\{\vb{I}_i\in\mathbb{R}^{H\times W\times 3}|i=1,\ldots,T\}$ featuring an RSP user, where $T$ represents the number of frames, and $H, W$ denote the image height and width. 
The task objective is to recover the 3D positions of both the available natural body joints $\vb{J}\in\mathbb{R}^{N_1\times 3}$ and points describing the RSP shape $\vb{P}\in\mathbb{R}^{N_2\times 3}$ for each image $\vb{I}\in\mathcal{V}$.
By capturing both natural body parts and the unique geometry of the RSP, our task aims to provide a foundation for comprehensive motion analysis, facilitating applications such as rehabilitation assessment and prosthesis design optimization.

\subsection{Evaluation Metrics}

\paragraph{Natural Body Joints}
We follow established practices in human body pose estimation research to compute the Mean Per Joint Position Error (\textbf{MPJPE}), which considers all available natural body joints $\bar{\vb{J}}$ and compares the ground truth positions with the estimated $\vb{J}$ in a root-aligned space. 
We further report the Procrustes-aligned MPJPE (\textbf{MPJPE-PA}), which eliminates global differences in scale, translation, and rotation, resulting in a more pose-focused error metric. 
Both metrics are reported in \texttt{mm}.

\paragraph{RSP Shape} 
We evaluate the RSP by first examining it within the body space.
This is essential for assessing the accurate interaction between the RSP and the body under the overall body configuration, which is crucial for prosthesis-aware pose estimation. 
Subsequently, we isolate the body pose estimation and focus on the local geometry of the RSP. 
We evenly sample the ground truth RSP surface $\bar{\mathcal{S}}$ to derive a point cloud $\bar{\vb{P}}$. 
Inspired by research in human object reconstruction~\cite{xie2024template}, we report bidirectional Chamfer Distance and its F-score between  $\bar{\vb{P}}$ and the estimated $\vb{P}$.

Specifically, we first translate both the estimated and ground truth RSPs by aligning them using the body pelvis positions. 
In this context, we report the bidirectional Chamfer Distance (\textbf{CD}) and its F-score at thresholds of \SI{100}{mm} and \SI{50}{mm} (\textbf{F@100}, \textbf{F@50}). Next, to focus on the local geometry, we align the average positions of $\vb{P},\bar{\vb{P}}$.
We then re-evaluate the Chamfer Distance (\textbf{CD-C}) and report the F-score at thresholds of \SI{50}{mm}, \SI{30}{mm} (\textbf{F-C@50}, \textbf{F-C@30}).

\begin{figure}[!t]
\centering
\includegraphics[width=.99\textwidth]{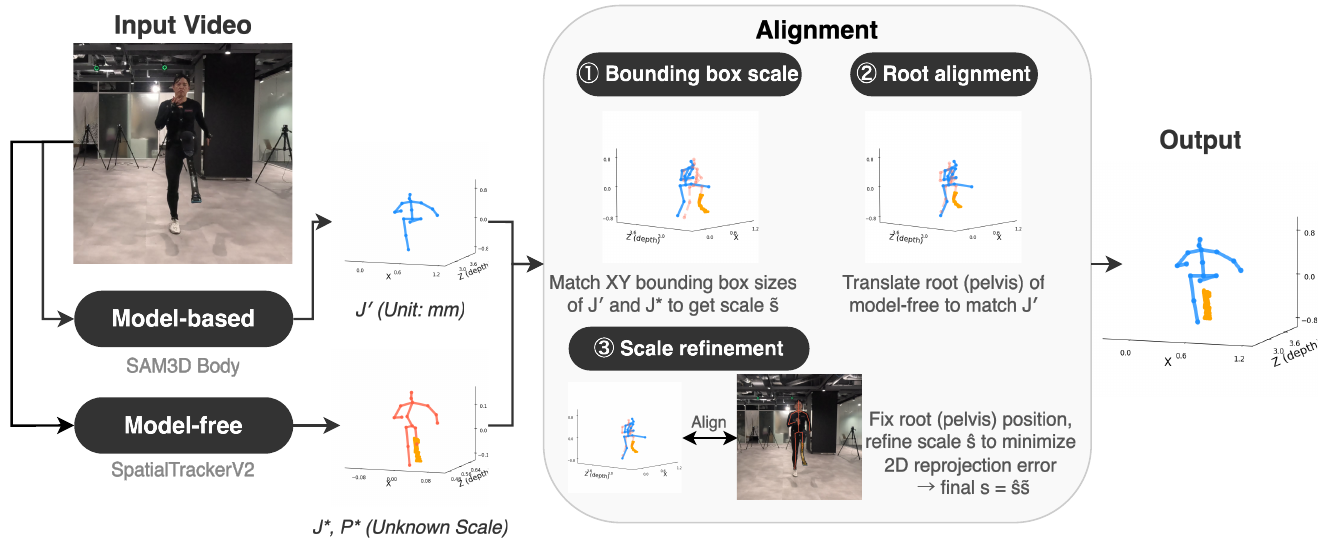}
\caption{Proposed hybrid baseline solution, which solves scale ambiguity to align the model-based and model-free solutions.}\label{fig:method}
\end{figure}

\subsection{Baseline Methods}\label{sec:eval_methods}
We evaluate all methods in a zero-shot setting, where pre-trained models are applied directly to our task without fine-tuning on our dataset. 
For each recorded action segment, we select one of the 16 GoPro cameras that offers moderate occlusion as the starting point for prosthesis-aware pose estimation.
We examine methods in two groups: model-based 3D human pose estimators and model-free reconstruction and tracking solutions.
We then establish the baseline for our task by integrating these two approaches.

\subsubsection*{Model-based}
The model-based 3D human pose estimator aims to recover the 3D positions of natural body joints $\vb{J}_{1:T}$ from the video observations $\mathcal{V}$.
Existing methods output joints covering the standard body structure~\cite{yang2026sam3dbody,shin2024wham,patel2025camerahmr,dwivedi2024tokenhmr,zhu2023motionbert}, while recent research enables amputation-aware body estimation~\cite{cho2025ajahr}.

We evaluate three representative model-based approaches. 
The first is SAM3D Body~\cite{yang2026sam3dbody}, a state-of-the-art method that regresses body joints and mesh directly from input images. 
The second is MotionBERT~\cite{zhu2023motionbert}, a lifting-based approach that converts 2D joint detections into 3D space. 
We assess MotionBERT using both 2D detections from AlphaPose~\cite{fang2022alphapose} and the 2D ground truth, with the latter providing an oracle analysis.
The third approach, AJAHR~\cite{cho2025ajahr}, differs from SAM3D Body and MotionBERT, which assume a standard set of body joints.
This is a recent prosthesis-aware solution that identifies amputations and outputs only the available body joints from images.

\subsubsection*{Model-free}
To recover both natural body joints $\vb{J}_{1:T}$ and RSP points $\vb{P}_{1:T}$, we apply model-free solutions that can handle arbitrary shapes.
Utilizing 2D joint detection and prosthesis-aware mask segmentation, these methods output $\vb{P}$ corresponding to the masked RSP points $\vb{p}\in\mathbb{R}^{N_2 \times 2}$ and assign 3D coordinates to $\vb{J}$ corresponding to detected 2D body joints $\vb{j}\in\mathbb{R}^{N_1 \times 2}$.

We refer to SpatialTrackerV2~\cite{xiao2025spatialtrackerv2} as the representative work, which facilitates efficient simultaneous 3D point tracking and scene reconstruction. 
This approach offers two solutions for our task: (1) \texttt{STv2-Pointmap}, which reads out surface points $\vb{J}_{1:T}, \vb{P}_{1:T}$ from the reconstructed scene point map by referencing the frame-wise 2D detections $\vb{j}_{1:T}, \vb{p}_{1:T}$;
(2) \texttt{STv2-Tracking}, which tracks the first frame detection $\vb{j}_1, \vb{p}_1$ in 3D space to obtain $\vb{J}_{1:T}, \vb{P}_{1:T}$ for the entire video.
We examine both solutions, employing AlphaPose~\cite{fang2022alphapose} and SAM2~\cite{ravi2025sam,ren2024grounding,ren2024grounded} to obtain $\vb{j}_{1:T}, \vb{p}_{1:T}$, respectively. 
We also report oracle results using the 2D ground truth. 

For implementation details, we obtain the detected RSP mask $\vb{p}$ by using SAM2 to track from the initial 2D ground truth bounding box of the RSP in the first frame, and we restart the tracking process every \SI{10}{sec}. 
For SpatialTrackerV2, we further divide the video into short chunks by setting $T=60$, which spans \SI{1}{sec} under the original 60 FPS recording.
Furthermore, when evaluating model-free solutions, to address the inherent scale ambiguity in monocular observations, we compare the bounding box size in the XY dimensions between the first frame $\vb{J}_1$ and its ground truth counterpart $\bar{\vb{J}}_1$.
This comparison allows us to recover the scale $s$, converting the outputs into \texttt{mm} measurements for evaluation.

\subsubsection*{Hybrid Baseline}
As illustrated in \cref{fig:method}, we establish the baseline for our proposed task by combining model-based and model-free solutions. 
Let the model-based outputs be denoted as $\vb{J}^\prime$, and the model-free outputs as $\vb{J}^\ast$ and $\vb{P}^\ast$. 
Given the groundtruth amputation site label, our alignment derives $\vb{J}$ directly from $\vb{J}^\prime$ by removing the hallucinated joints, while $\vb{P}$ is obtained by resolving the scale ambiguity in $\vb{P}^\ast$ and aligning it with the position of $\vb{J}^\prime$
\begin{equation}
\vb{J}=\vb{J}^\prime, \quad \vb{P} = s(\vb{P}^\ast-\vb*{r}^\ast)+\vb*{r}^\prime,\label{eq:hybrid}
\end{equation}
where $s\in\mathbb{R}$ is the scale factor and $\vb*{r}^\ast,\vb*{r}^\prime$ are the root joint of $\vb{J}^\ast, \vb{J}^\prime$.

We obtain $s$ through a three-step process to stabilize the computation. 
We begin by comparing the bounding box sizes in the XY-dimensions between $\vb{J}^\prime$ and $\vb{J}^\ast$. 
This comparison yields the scale factor $\tilde{s} \in \mathbb{R}$, which transforms $\vb{J}^\ast$ to a scale comparable with $\vb{J}^\prime$. 
Next, we align the root (\ie, pelvis) of the model-free outputs to that of the model-based $\vb{J}^\prime$, addressing the depth discrepancy between $\vb{J}^\prime$ and $\tilde{s}\vb{J}^\ast$ caused by inaccurate depth estimation. 
The transformation is given by:
\begin{equation}
\hat{\vb{J}}^\ast = \tilde{s}(\vb{J}^\ast-\vb*{r}^\ast)+\vb*{r}^\prime, \quad \hat{\vb{P}}^\ast = \tilde{s}(\vb{P}^\ast-\vb*{r}^\ast)+\vb*{r}^\prime.
\end{equation}
Finally, we refine the scaling factor to ensure consistent 2D projection between $\vb{P}^\ast$ and $\vb{P}$:
\begin{equation}
\hat{s} = \arg\min_{\hat{s}>0} \left\| \pi\left(\hat{s}\left(\begin{bmatrix} \hat{\vb{J}}^\ast \\ \hat{\vb{P}}^\ast \end{bmatrix} - \vb*{r}^\prime\right) + \vb*{r}^\prime\right) - \pi\left(\begin{bmatrix} \vb{J}^\ast \\ \vb{P}^\ast \end{bmatrix}\right) \right\|,
\end{equation}
where $\pi(.)$ denotes the perspective projection using the recording intrinsics. 
We therefore obtain $s = \hat{s}\tilde{s}$.

We use SAM3D Body~\cite{yang2026sam3dbody} and STv2-Pointmap~\cite{xiao2025spatialtrackerv2} as baseline methods due to their superior performance.
Similar to the evaluation of model-free methods, we examine performance using either 2D detection data or 2D GT counterparts when implementing STv2-Pointmap.

\section{Results}

\begin{figure}[!t]
\centering
\includegraphics[width=.99\textwidth]{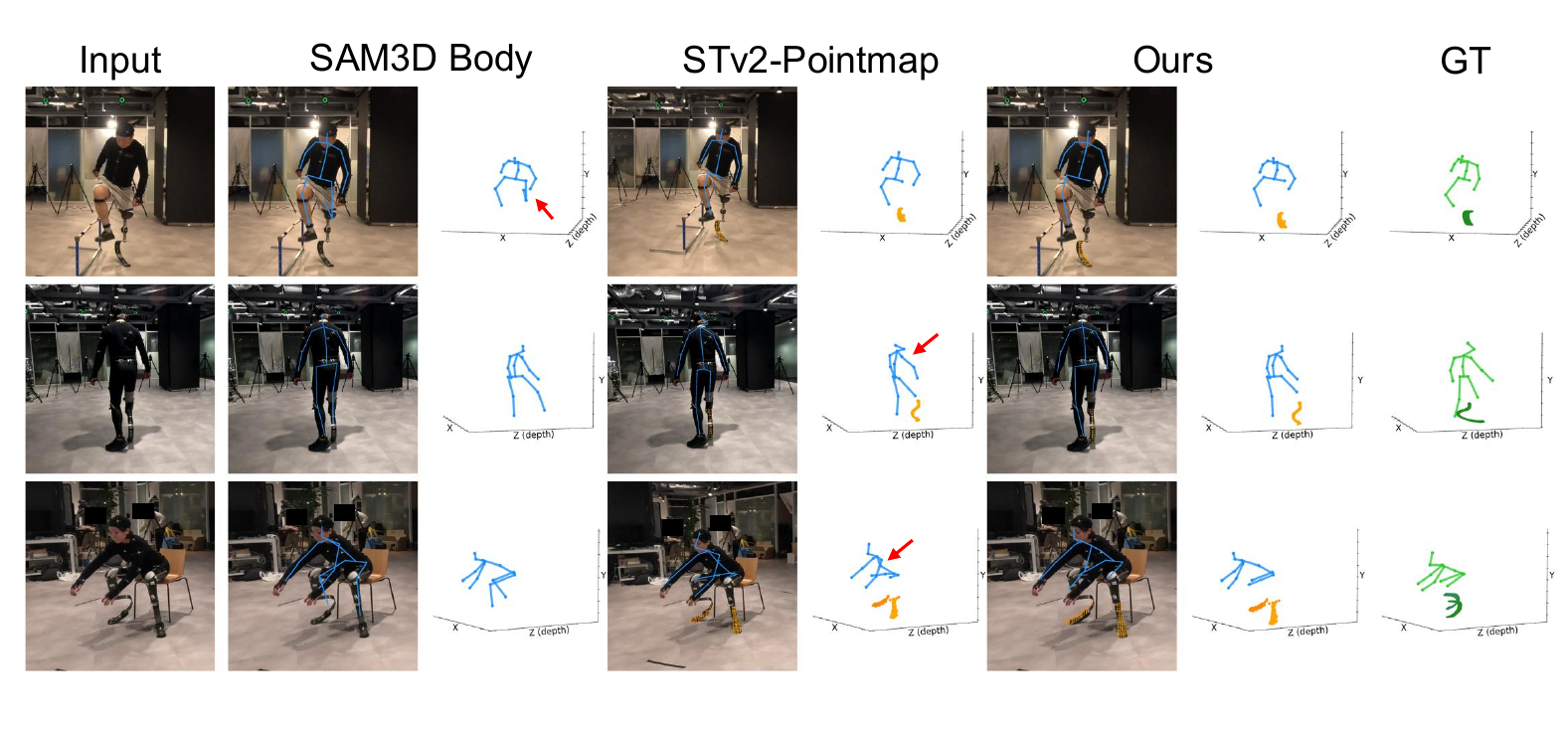}
\caption{Qualitative comparisons with SAM3D Body, SpatialTrackerV2-Pointmap and the proposed hybrid alignment baseline.}\label{fig:quali}
\end{figure}

\cref{tab:baseline_metrics} and \cref{fig:quali} summarize the quantitative and qualitative results of all evaluated methods on \datasetname.
We begin by discussing the specific limitations of both model-based and model-free approaches. 
Following this, we analyze how our hybrid baseline effectively addresses these limitations and further examine performance variations across different amputation conditions.

\begin{table}[!t]
\centering
\caption{Quantitative results on the collected dataset. 
For model-free solutions, outputs are aligned with the GT to resolve scale ambiguity.}
\label{tab:baseline_metrics}
\resizebox{\linewidth}{!}{%
\begin{tabular}{c|c|cc|cccccc}
\toprule
 & Method & MPJPE $\downarrow$ & MPJPE-PA $\downarrow$  & CD $\downarrow$ & F@100 $\uparrow$ & F@50 $\uparrow$ & CD-C $\downarrow$  & F-C@50 $\uparrow$ & F-C@30 $\uparrow$ \\
\midrule
\multirow{4}{*}{Model-Based} 
& AJAHR~\cite{cho2025ajahr} & 125.06 & 90.85  & - & - & - & - & - & -  \\
& SAM3D Body~\cite{yang2026sam3dbody} & \textbf{78.35} & \textbf{50.89} & - & - & - & - & - & - \\
& MotionBERT~\cite{zhu2023motionbert} (w/ 2D Det.) & 115.06 & 64.94  & - & - & - & - & - & - \\
& MotionBERT~\cite{zhu2023motionbert} (w/ 2D GT) & 112.41 & 60.03  & - & - & - & - & - & - \\
\hline
\multirow{2}{*}{\makecell{Model-Free\\ (w/ 2D Detection)}} 
& STv2-Tracking~\cite{xiao2025spatialtrackerv2} & 174.84 & 125.45 & 299.59 & 40.31 & 15.32 & 122.42 & 48.32 & 22.49 \\
& STv2-Pointmap~\cite{xiao2025spatialtrackerv2} & 159.64 & 113.87 & \textbf{277.15} & 45.44 & 18.82 & 115.00 & \textbf{50.60} & 24.33 \\
\hline
\makecell{Hybrid Alignment \\ (w/ 2D Detection)}
& \makecell{SAM3D Body~\cite{yang2026sam3dbody} \\ + STv2-Pointmap~\cite{xiao2025spatialtrackerv2}} & 
\textbf{78.35} & \textbf{50.89} & 288.44 & \textbf{45.61} & \textbf{19.92} & 119.60 & 50.45 & \textbf{25.15} \\
\midrule
\multirow{2}{*}{\makecell{Model-Free\\ (w/ 2D GT)}} & STv2-Tracking~\cite{xiao2025spatialtrackerv2} & 141.14 & 109.71 & 273.98 & 46.17 & 19.30 & 124.67 & 50.99 & 24.16 \\
& STv2-Pointmap~\cite{xiao2025spatialtrackerv2} & 119.59 & 93.94 & \textbf{238.71} & \textbf{53.81} & \textbf{25.74} & \textbf{115.63} & 54.60 & 26.72 \\
\midrule
\makecell{Hybrid Alignment \\ (w/ 2D GT)}
& \makecell{SAM3D Body~\cite{yang2026sam3dbody} \\ + STv2-Pointmap~\cite{xiao2025spatialtrackerv2}} & \textbf{78.35} & \textbf{50.89} & 251.15 & 50.72 & 22.41 & 117.23 & \textbf{54.94} & \textbf{27.52} \\
\bottomrule
\end{tabular}}
\end{table}

\subsection{Model-based and Model-free Solutions} 

\begin{figure}[!t]
\centering
\includegraphics[width=.99\textwidth]{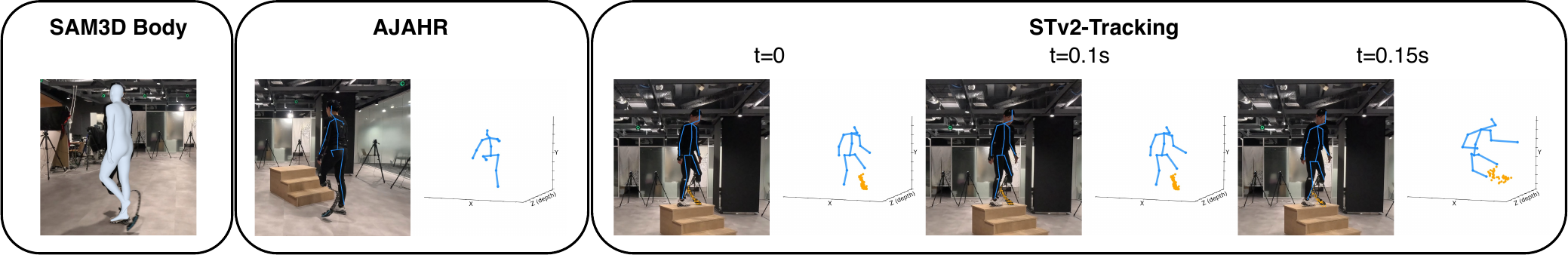}
\caption{Failure cases of SAM3D Body, AJAHR, and STv2-Tracking. 
SAM3D Body hallucinates amputated-side meshes that serve fundamentally different biomechanical functions from RSPs and have drastically different geometries. 
AJAHR shows reduced accuracy due to difficulties in classifying amputation under occlusion. 
STv2-Tracking struggles with the thin geometry of RSPs and textureless clothing.}\label{fig:failure}
\end{figure}

Model-based solutions are fundamentally unable to capture the 3D geometry of prostheses, as their outputs are limited to a predefined set of body joints based on human anatomy,
as exemplified by the qualitative results of SAM3D Body in \cref{fig:quali,fig:failure}.
Although they leverage learned spatial relationships among joints to estimate the positions of missing body joints, this prediction often fails to align with the observed prosthesis due to the domain gap caused by the appearance difference between prosthetic and natural limbs.

Conversely, model-free solutions are not limited to predefined object categories and can thus handle prosthesis estimation.
However, they have limited ability to use kinematic and geometric priors of the human body.
As demonstrated in~\cref{tab:baseline_metrics} and illustrated in~\cref{fig:quali}, this limitation hinders reliable estimation under occlusion, leading to worse accuracy for natural body joints.

Among model-based approaches, SAM3D Body outperforms MotionBERT and AJAHR in estimating body joints.
We attribute this to SAM3D Body's large and diverse training data, along with its ability to use richer appearance information from input images.
As shown in \cref{fig:failure}, while AJAHR accounts for amputations in its output, it struggles to correctly classify amputation status under occlusion, and such misclassifications of natural limbs as amputated penalize its accuracy.

For SpatialTrackerV2, retrieving outputs from the reconstructed scene point map yields better results than those obtained through tracking.
As exemplified in \cref{fig:failure}, frequent tracking loss is observed, primarily due to the lack of texture on clothing and the thin geometry of the RSP.
This problem is more severe for model-free methods because they cannot incorporate prior knowledge of RSP shape or body kinematics.
Furthermore, comparing results with 2D detection inputs against those with 2D ground truth reveals that point selection based on 2D detections is a significant bottleneck for model-free solutions: even with a fixed 3D scene reconstruction, incorrect 2D detections cause wrong points to be sampled from the point map, directly hurting accuracy. 
This suggests that improving 2D detection and segmentation for prosthesis users is a promising direction for future work.

\begin{table}[!t]
\centering
\caption{Comparison among users with different amputation types. Both methods leverage 2D detection for prediction.}\label{tab:comparison_amputation}
\resizebox{0.99\textwidth}{!}{
\begin{tabular}{c|c|cc|ccc|ccc}
\midrule
Amputation Type & Method & MPJPE $\downarrow$ & MPJPE-PA $\downarrow$  & CD $\downarrow$ & F@100 $\uparrow$ & F@50 $\uparrow$ & CD-C $\downarrow$  & F-C@50 $\uparrow$ & F-C@30 $\uparrow$ \\
\midrule
\multirow{2}{*}{Bilateral} 
& STv2-Pointmap & 155.60 & 98.89 & \textbf{294.21} & \textbf{45.88} & \textbf{19.56} & \textbf{106.80} & \textbf{54.60} & 25.97 \\
& Hybrid Alignment & \textbf{91.91} & \textbf{51.32} & 342.22 & 37.41 & 13.80 & 113.32 & 54.25 & \textbf{27.35} \\
\midrule
\multirow{2}{*}{Unilateral} 
& STv2-Pointmap  & 162.02 & 122.71 & 267.07 & 45.19 & 18.38 & \textbf{119.84} & \textbf{48.25} & 23.36 \\
& Hybrid Alignment & \textbf{70.35} & \textbf{50.63} & \textbf{256.64} & \textbf{50.45} & \textbf{23.53} & 123.31 & 48.21 & \textbf{23.85} \\

\midrule
\end{tabular}}
\end{table}

\subsection{Hybrid Baseline}

The two limitations identified above, namely the inability of model-based methods to capture prosthesis geometry and the lack of body priors in model-free methods, directly motivate us to establish the baseline with a hybrid approach.
Results in \cref{tab:baseline_metrics} confirm that combining the two components addresses both limitations: the model-based component provides accurate body joint estimation, while the model-free component recovers the RSP geometry. 
Qualitative results in \cref{fig:quali} also show that our hybrid baseline achieves robust estimation of natural body joints, even when joints are absent or occluded. 
Additionally, it captures the 3D geometry and pose of the RSP.

As we directly adopt the estimated body joints from the model-based component, we further discuss by focusing on the performance of RSP estimation to evaluate the effectiveness of scale alignment. 
Our results demonstrate overall comparable performance to model-free solutions that resolve scale ambiguity by referring to the 3D ground truth, which verifies the effectiveness of our proposed scale alignment.
However, we find that accurate estimation in the root-aligned space remains challenging, particularly reflected by the bidirectional Chamfer Distance. 
We attribute this to the sensitivity of alignment to inaccurate depth estimation, highlighting that achieving global alignment between body parts and RSP is a key challenge for future work.

Moreover, as reported in \cref{tab:comparison_amputation}, participants with bilateral amputations generally show higher estimation errors compared to those with unilateral amputations, largely due to the increased difficulty in alignment when both legs are replaced by prostheses. 
On one hand, the model-based component has fewer intact joints available as anchor points for below-knee scale alignment, making it less stable in capturing the relative positions between both RSPs and the body. 
On the other hand, higher errors in each component for participants with bilateral amputations further complicate accurate scale recovery.
We further discuss per-participant results in the supplementary materials.

\section{Conclusion}

Recovering 3D human body motion from video has important applications in rehabilitation assessment and sports performance evaluation, yet existing methods have not been designed for prosthesis users.
In this work, we introduce the task of prosthesis-aware 3D human pose estimation and collect \datasetname, the first 3D dataset of RSP users, covering a range of daily-life and exercise actions from participants with diverse amputation conditions.
Our evaluation confirms that model-based methods cannot represent RSP geometry, while model-free methods lack body kinematic priors and are unreliable under occlusion.
Our hybrid baseline combines the strengths of both approaches and establishes a starting point for future research on this task.

\paragraph{Limitation and Future Work}
The current dataset scale is insufficient for reliable training. 
Dataset expansion and extension to marker-less, in-the-wild settings remain important future directions. 
The zero-shot hybrid baseline serves only as a reference, and more advanced solutions could be explored, such as leveraging synthetic data or incorporating biomechanical constraints and RSP consistency into an optimization framework. 
Amputation site labels are currently assigned manually; integrating the temporal cue for automatic prediction remains an open direction. 
Finally, while the unsaturated baselines demonstrate task difficulty, a more comprehensive evaluation of RSP shapes and poses and stricter error thresholds are left for future work.

\paragraph{Acknowledgment} 
This research is supported by JSPS KAKENHI Grant Number JP25K03134, Toyota Foundation Grant Number D24-ST-0030, JST ASPIRE Grant Number JPMJAP2303, and The Telecommunications Advancement Foundation.
The authors are also grateful to Atsuro Okino (OSPO), Motohiko Takahashi (Step4ward), and Hideto Naito (Xiborg) for helpful discussions and support.

\appendix
\section{Overview}
In this supplementary material, we present detailed action descriptions and per-participant baseline results.
Additionally, we include a video titled \texttt{supp.mp4}, which is compatible with most media players. 
This video demonstrates the recorded actions, exemplifies the various motion patterns among participants, and offers qualitative comparisons, thereby supporting the discussion in our main text.

\section{Dataset Actions}

We demonstrate the detailed actions we ask participants to perform.
To ensure safety, we made slight adjustments to the treadmill speed and the number of repetitions based on the participant's stamina and ability during the recording. 
We encourage participants to perform the actions to the best of their ability, even if they find it challenging to execute them in the standard manner. 
The following sections describe the actions, with demos available in the supplementary video.

\paragraph{A1: Treadmill Walking}
Begin walking at a speed of 1.0 km/h, and increase the speed by 1 km/h every 15 seconds, up to a maximum of 4 km/h or the participant's maximum comfortable speed. 
Maintain this speed for 30 seconds. 
Then, decrease the speed by 1 km/h every 15 seconds until returning to 1.0 km/h.

\paragraph{A2: Treadmill Jogging}
Start jogging at a speed of 4.0 km/h, and increase the speed by 1 km/h every 15 seconds, up to a maximum of 7 km/h or the participant's maximum comfortable speed. 
Maintain this speed for 30 seconds. 
Then, decrease the speed by 1 km/h every 15 seconds until returning to 4.0 km/h.

\paragraph{A3: Incline Walking}
Maintain a fixed speed while setting the incline to 5\% or 10\% and walk for 30 seconds.

\paragraph{B1: Sit-to-Stand (Stairs)}
Sit on the third step of the stairs and stand up.

\paragraph{B2: Stair Ascent/Descent}
Go up and down the stairs.

\paragraph{C1: Stepping Over Low Hurdles}
Continuously step over mini hurdles, performing the exercise while facing both forward and sideways.

\paragraph{C2: Stepping Over High Hurdles}
Step over hurdles approximately at knee height. Perform the exercise by facing forward and leading with either the right or left foot. 
Additionally, perform round trips while facing sideways.

\paragraph{D1: Free Walking}
Walk along a straight line or circle.

\paragraph{D2: Free Jogging}
Jog lightly along a straight line or circle.

\paragraph{D3: Hip Rotation}
Rotate the hips while standing. 
Perform both clockwise and counterclockwise.

\paragraph{D4: Sit-to-Stand (Chair)}
Sit on the chair and then stand up.

\paragraph{D5: Side Steps}
Move sideways while keeping the body facing forward, then return to the original position. 

\paragraph{D6: Squats}
From a standing position, bend the knees and lower the hips while pulling the buttocks backward, then stand up. 

\paragraph{D7: Single-leg T-Balance}
Stand on the non-amputated leg and lean the upper body forward. 
Extend the upper limbs sideways to form a T-shape with the body and hold the position for 3 seconds. 

\paragraph{D8: Rotation in Place}
Pivot on one foot and rotate in place. 
Use either the right or left foot as the axis, performing rotations both clockwise and counterclockwise.

\paragraph{D9: Sit-to-Stand (Floor)}
Sit down on the floor and then stand up.

\paragraph{E1: High Knees}
Perform a rapid jogging motion in place, lifting each knee to hip height with each step.

\paragraph{E2: Crouching Start}
From a standing position, lower the posture, place both hands on the ground, place one knee up, and shift the weight forward, preparing for the start signal, then return to a standing position.

\paragraph{E3: Vertical Jump}
From a stationary position, jump straight up as high as possible, using maximum effort for each jump. 
Avoid taking a running start and focus on explosive power.

\paragraph{E4: Straight Leg Jump (Scissors)}
With knees extended, jump by alternately extending each foot forward.

\paragraph{E5: Toe Jump (Pogo)}
Jump lightly in place with both feet together and knees slightly bent, maintaining a continuous bouncing motion.

\section{Result for Each Participant}

We report the per-participant results for the proposed hybrid alignment in \cref{tab:per_subject}, where 2D detection is referenced during the inference stage. 
Two observations emerge from this per-participant analysis:

First, compared to participants with unilateral amputation, those with bilateral amputation (\ie, \texttt{P5}, \texttt{P6}) exhibit higher joint error in terms of MPJPE. 
This increased error may be attributed to the fact that both lower limbs are replaced by running-specific prostheses (RSPs), which increase the difficulty for model-based methods to generalize well.
As a result, the estimated positions of natural body joints become less accurate. 
The imprecise estimation of natural body joints, coupled with the reduced number of intact joints available for below-knee scale alignment, makes the subsequent hybrid alignment less reliable. 
This leads to inaccuracies when evaluating prosthetic blades in the body space (\ie, CD and its F-score), which is consistent with the discussion in the main text.

\begin{figure}[!t]
\centering
\includegraphics[width=.99\textwidth]{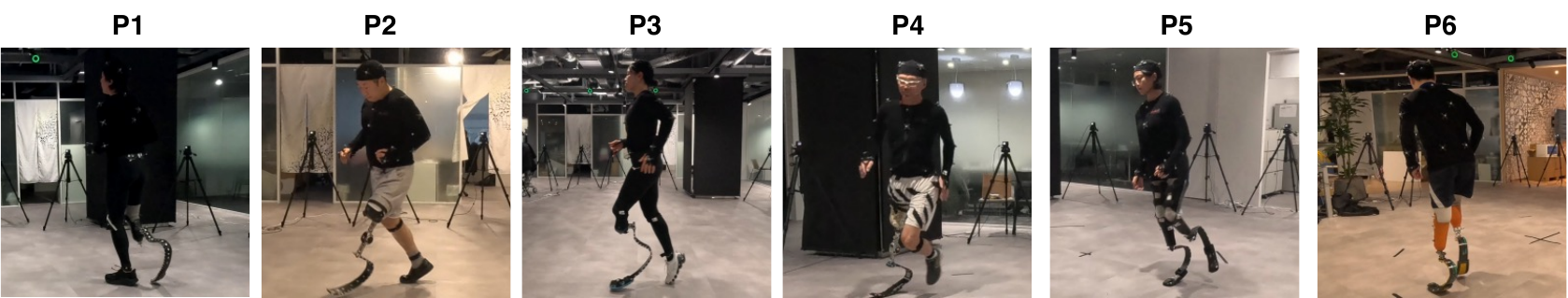}
\caption{Sample image of each participant.}\label{fig:participant}
\end{figure}

Second, when evaluating the local geometry of the RSP estimation by aligning the average positions of the estimation with the corresponding ground truth, we find significantly worse results for \texttt{P1} and \texttt{P3}, in terms of CD-C and its F-score. 
We attribute this to the fact that \texttt{P1} and \texttt{P3} are elite athletes using RSPs with more pronounced curvature (see \cref{fig:participant}), which complicates the ability of model-free methods to capture the prior knowledge of RSP geometry for accurate depth estimation.

\begin{table}[!tp]
\centering
\caption{Per-participant results for the proposed \textit{Hybrid Alignment (w/ 2D Detection)} baseline.}\label{tab:per_subject}
\resizebox{\linewidth}{!}{
\begin{tabular}{cc|cc|ccc|ccc}
\hline
\textbf{ID} & \textbf{Amputation Site} & \textbf{MPJPE} $\downarrow$ & \textbf{MPJPE-PA} $\downarrow$ & \textbf{CD} $\downarrow$ & \textbf{F@100} $\uparrow$ & \textbf{F@50} $\uparrow$ & \textbf{CD-C} $\downarrow$ & \textbf{F-C@50} $\uparrow$ & \textbf{F-C@30} $\uparrow$ \\
\hline
P1 & Below-knee (Right) & 61.91 & 49.71 & 255.20 & 46.37 & 21.43 & 131.22 & 44.47 & 22.05 \\
P2 & Above-knee (Left) & 60.30 & 42.74 & 283.87 & 46.18 & 19.18 & 114.23 & 52.62 & 25.64 \\
P3 & Below-knee (Left) & 83.27 & 52.16 & 247.49 & 51.64 & 24.07 & 132.04 & 44.66 & 23.10 \\
P4 & Above-knee (Right) & 73.30 & 55.92 & 248.96 & 57.52 & 28.89 & 109.44 & 54.05 & 25.74 \\
P5 & Below-knee (Bilateral) & 102.20 & 55.07 & 319.29 & 37.77 & 14.47 & 115.57 & 52.12 & 24.76 \\
P6 & Below-knee (Bilateral) & 83.34 & 48.19 & 361.34 & 37.11 & 13.25 & 111.45 & 56.03 & 29.51 \\
\hline
\end{tabular}}
\end{table}

\bibliographystyle{splncs04}

\begin{thebibliography}{10}
\providecommand{\url}[1]{\texttt{#1}}
\providecommand{\urlprefix}{URL }
\providecommand{\doi}[1]{https://doi.org/#1}

\bibitem{bregler2000recovering}
Bregler, C., Hertzmann, A., Biermann, H.: Recovering non-rigid 3d shape from image streams. In: Proceedings IEEE Conference on Computer Vision and Pattern Recognition. CVPR 2000 (Cat. No. PR00662). vol.~2, pp. 690--696. IEEE (2000)

\bibitem{chen2025easi3r}
Chen, X., Chen, Y., Xiu, Y., Geiger, A., Chen, A.: Easi3r: Estimating disentangled motion from dust3r without training. In: Proceedings of the IEEE/CVF International Conference on Computer Vision. pp. 9158--9168 (2025)

\bibitem{chen2026sam}
Chen, X., Chu, F.J., Gleize, P., Liang, K.J., Sax, A., Tang, H., Wang, W., Guo, M., Hardin, T., Li, X., et~al.: Sam 3d: 3dfy anything in images. In: Proceedings of the IEEE/CVF Conference on Computer Vision and Pattern Recognition. pp. 7220--7232 (2026)

\bibitem{cho2025ajahr}
Cho, H., Choi, G., Choi, J.: Ajahr: Amputated joint aware 3d human mesh recovery. In: Proceedings of the IEEE/CVF International Conference on Computer Vision. pp. 7925--7935 (2025)

\bibitem{cseke2025pico}
Cseke, A., Tripathi, S., Dwivedi, S.K., Lakshmipathy, A.S., Chatterjee, A., Black, M.J., Tzionas, D.: Pico: Reconstructing 3d people in contact with objects. In: Proceedings of the Computer Vision and Pattern Recognition Conference. pp. 1783--1794 (2025)

\bibitem{dabhi20243d}
Dabhi, M., Jeni, L.A., Lucey, S.: 3d-lfm: Lifting foundation model. In: Proceedings of the IEEE/CVF Conference on Computer Vision and Pattern Recognition. pp. 10466--10475 (2024)

\bibitem{dai2014simple}
Dai, Y., Li, H., He, M.: A simple prior-free method for non-rigid structure-from-motion factorization. International Journal of Computer Vision  \textbf{107}(2),  101--122 (2014)

\bibitem{du2026inclusivevidpose}
Du, H., Ying, J., Wang, S., Li, X., Zhang, K., Yu, X.: Inclusivevidpose: Bridging the pose estimation gap for individuals with limb deficiencies in video-based motion. In: International Conference on Learning Representations (ICLR) (2026)

\bibitem{dwivedi2024tokenhmr}
Dwivedi, S.K., Sun, Y., Patel, P., Feng, Y., Black, M.J.: Tokenhmr: Advancing human mesh recovery with a tokenized pose representation. In: Proceedings of the IEEE/CVF Conference on Computer Vision and Pattern Recognition. pp. 1323--1333 (2024)

\bibitem{fang2022alphapose}
Fang, H.S., Li, J., Tang, H., Xu, C., Zhu, H., Xiu, Y., Li, Y.L., Lu, C.: Alphapose: Whole-body regional multi-person pose estimation and tracking in real-time. IEEE transactions on pattern analysis and machine intelligence  \textbf{45}(6),  7157--7173 (2022)

\bibitem{feng2025st4rtrack}
Feng, H., Zhang, J., Wang, Q., Ye, Y., Yu, P., Black, M.J., Darrell, T., Kanazawa, A.: St4rtrack: Simultaneous 4d reconstruction and tracking in the world. In: Proceedings of the IEEE/CVF International Conference on Computer Vision. pp. 8503--8513 (2025)

\bibitem{MHR:2025}
Ferguson, A., Osman, A.A.A., Bescos, B., Stoll, C., Twigg, C., Lassner, C., Otte, D., Vignola, E., Prada, F., Bogo, F., Santesteban, I., Romero, J., Zarate, J., Lee, J., Park, J., Yang, J., Doublestein, J., Venkateshan, K., Kitani, K., Kavan, L., Farra, M.D., Hu, M., Cioffi, M., Fabris, M., Ranieri, M., Modarres, M., Kadlecek, P., Khirodkar, R., Abdrashitov, R., Prévost, R., Rajbhandari, R., Mallet, R., Pearsall, R., Kao, S., Kumar, S., Parrish, S., Yu, S.I., Saito, S., Shiratori, T., Wang, T.L., Tung, T., Xu, Y., Dong, Y., Chen, Y., Xu, Y., Ye, Y., Jiang, Z.: Mhr: Momentum human rig. arXiv preprint arXiv:2511.15586  (2025)

\bibitem{garrido2014automatic}
Garrido-Jurado, S., Mu{\~n}oz-Salinas, R., Madrid-Cuevas, F.J., Mar{\'\i}n-Jim{\'e}nez, M.J.: Automatic generation and detection of highly reliable fiducial markers under occlusion. Pattern Recognition  \textbf{47}(6),  2280--2292 (2014)

\bibitem{goel2023humans}
Goel, S., Pavlakos, G., Rajasegaran, J., Kanazawa, A., Malik, J.: Humans in 4d: Reconstructing and tracking humans with transformers. In: Proceedings of the IEEE/CVF International Conference on Computer Vision. pp. 14783--14794 (2023)

\bibitem{grauman2024ego}
Grauman, K., Westbury, A., Torresani, L., Kitani, K., Malik, J., Afouras, T., Ashutosh, K., Baiyya, V., Bansal, S., Boote, B., et~al.: Ego-exo4d: Understanding skilled human activity from first-and third-person perspectives. In: Proceedings of the IEEE/CVF Conference on Computer Vision and Pattern Recognition. pp. 19383--19400 (2024)

\bibitem{ionescu2013human3}
Ionescu, C., Papava, D., Olaru, V., Sminchisescu, C.: Human3. 6m: Large scale datasets and predictive methods for 3d human sensing in natural environments. IEEE transactions on pattern analysis and machine intelligence  \textbf{36}(7),  1325--1339 (2013)

\bibitem{kanazawa2018end}
Kanazawa, A., Black, M.J., Jacobs, D.W., Malik, J.: End-to-end recovery of human shape and pose. In: Proceedings of the IEEE conference on computer vision and pattern recognition. pp. 7122--7131 (2018)

\bibitem{kocabas2020vibe}
Kocabas, M., Athanasiou, N., Black, M.J.: Vibe: Video inference for human body pose and shape estimation. In: Proceedings of the IEEE/CVF conference on computer vision and pattern recognition. pp. 5253--5263 (2020)

\bibitem{kolotouros2019learning}
Kolotouros, N., Pavlakos, G., Black, M.J., Daniilidis, K.: Learning to reconstruct 3d human pose and shape via model-fitting in the loop. In: Proceedings of the IEEE/CVF international conference on computer vision. pp. 2252--2261 (2019)

\bibitem{kumar2020non}
Kumar, S.: Non-rigid structure from motion: Prior-free factorization method revisited. In: Proceedings of the IEEE/CVF Winter Conference on Applications of Computer Vision. pp. 51--60 (2020)

\bibitem{li2025genmo}
Li, J., Cao, J., Zhang, H., Rempe, D., Kautz, J., Iqbal, U., Yuan, Y.: Genmo: A generalist model for human motion. In: Proceedings of the IEEE/CVF International Conference on Computer Vision. pp. 11766--11776 (2025)

\bibitem{SMPL:2015}
Loper, M., Mahmood, N., Romero, J., Pons-Moll, G., Black, M.J.: {SMPL}: A skinned multi-person linear model. ACM Trans. Graphics (Proc. SIGGRAPH Asia)  \textbf{34}(6),  248:1--248:16 (Oct 2015)

\bibitem{martinez2017simple}
Martinez, J., Hossain, R., Romero, J., Little, J.J.: A simple yet effective baseline for 3d human pose estimation. In: Proceedings of the IEEE international conference on computer vision. pp. 2640--2649 (2017)

\bibitem{mehta2017monocular}
Mehta, D., Rhodin, H., Casas, D., Fua, P., Sotnychenko, O., Xu, W., Theobalt, C.: Monocular 3d human pose estimation in the wild using improved cnn supervision. In: 2017 international conference on 3D vision (3DV). pp. 506--516. IEEE (2017)

\bibitem{newell2016stacked}
Newell, A., Yang, K., Deng, J.: Stacked hourglass networks for human pose estimation. In: European conference on computer vision. pp. 483--499. Springer (2016)

\bibitem{oquab2024dinov2}
Oquab, M., Darcet, T., Moutakanni, T., Vo, H., Szafraniec, M., Khalidov, V., Fernandez, P., Haziza, D., Massa, F., El-Nouby, A., et~al.: Dinov2: Learning robust visual features without supervision. Transactions on Machine Learning Research Journal  (2024)

\bibitem{patel2025camerahmr}
Patel, P., Black, M.J.: Camerahmr: Aligning people with perspective. In: 2025 International Conference on 3D Vision (3DV). pp. 1562--1571. IEEE (2025)

\bibitem{ravi2025sam}
Ravi, N., Gabeur, V., Hu, Y.T., Hu, R., Ryali, C., Ma, T., Khedr, H., R{\"a}dle, R., Rolland, C., Gustafson, L., et~al.: Sam 2: Segment anything in images and videos. In: International Conference on Learning Representations. vol.~2025, pp. 28085--28128 (2025)

\bibitem{ren2024grounding}
Ren, T., Jiang, Q., Liu, S., Zeng, Z., Liu, W., Gao, H., Huang, H., Ma, Z., Jiang, X., Chen, Y., Xiong, Y., Zhang, H., Li, F., Tang, P., Yu, K., Zhang, L.: Grounding dino 1.5: Advance the "edge" of open-set object detection. arXiv preprint arXiv:2405.10300  (2024)

\bibitem{ren2024grounded}
Ren, T., Liu, S., Zeng, A., Lin, J., Li, K., Cao, H., Chen, J., Huang, X., Chen, Y., Yan, F., Zeng, Z., Zhang, H., Li, F., Yang, J., Li, H., Jiang, Q., Zhang, L.: Grounded sam: Assembling open-world models for diverse visual tasks. arXiv preprint arXiv:2401.14159  (2024)

\bibitem{shin2024wham}
Shin, S., Kim, J., Halilaj, E., Black, M.J.: Wham: Reconstructing world-grounded humans with accurate 3d motion. In: Proceedings of the IEEE/CVF Conference on Computer Vision and Pattern Recognition. pp. 2070--2080 (2024)

\bibitem{sun2019deep}
Sun, K., Xiao, B., Liu, D., Wang, J.: Deep high-resolution representation learning for human pose estimation. In: Proceedings of the IEEE/CVF conference on computer vision and pattern recognition. pp. 5693--5703 (2019)

\bibitem{von2018recovering}
Von~Marcard, T., Henschel, R., Black, M.J., Rosenhahn, B., Pons-Moll, G.: Recovering accurate 3d human pose in the wild using imus and a moving camera. In: Proceedings of the European conference on computer vision (ECCV). pp. 601--617 (2018)

\bibitem{wang2025vggt}
Wang, J., Chen, M., Karaev, N., Vedaldi, A., Rupprecht, C., Novotny, D.: Vggt: Visual geometry grounded transformer. In: Proceedings of the Computer Vision and Pattern Recognition Conference. pp. 5294--5306 (2025)

\bibitem{wang2025shape}
Wang, Q., Ye, V., Gao, H., Zeng, W., Austin, J., Li, Z., Kanazawa, A.: Shape of motion: 4d reconstruction from a single video. In: Proceedings of the IEEE/CVF International Conference on Computer Vision. pp. 9660--9672 (2025)

\bibitem{wang2025c4d}
Wang, S., Jiang, Z., Yang, X., Wang, X.: C4d: 4d made from 3d through dual correspondences. In: Proceedings of the IEEE/CVF International Conference on Computer Vision. pp. 7570--7580 (2025)

\bibitem{wang2024dust3r}
Wang, S., Leroy, V., Cabon, Y., Chidlovskii, B., Revaud, J.: Dust3r: Geometric 3d vision made easy. In: Proceedings of the IEEE/CVF conference on computer vision and pattern recognition. pp. 20697--20709 (2024)

\bibitem{wen2025reconstructing}
Wen, B., Huang, D., Zhang, Z., Zhou, J., Deng, J., Gong, J., Chen, Y., Ma, L., Li, Y.L.: Reconstructing in-the-wild open-vocabulary human-object interactions. In: Proceedings of the Computer Vision and Pattern Recognition Conference. pp. 17426--17436 (2025)

\bibitem{xiao2025spatialtrackerv2}
Xiao, Y., Wang, J., Xue, N., Karaev, N., Makarov, Y., Kang, B., Zhu, X., Bao, H., Shen, Y., Zhou, X.: Spatialtrackerv2: Advancing 3d point tracking with explicit camera motion. In: Proceedings of the IEEE/CVF International Conference on Computer Vision. pp. 6726--6737 (2025)

\bibitem{xiao2024spatialtracker}
Xiao, Y., Wang, Q., Zhang, S., Xue, N., Peng, S., Shen, Y., Zhou, X.: Spatialtracker: Tracking any 2d pixels in 3d space. In: Proceedings of the IEEE/CVF Conference on Computer Vision and Pattern Recognition. pp. 20406--20417 (2024)

\bibitem{xie2024template}
Xie, X., Bhatnagar, B.L., Lenssen, J.E., Pons-Moll, G.: Template free reconstruction of human-object interaction with procedural interaction generation. In: Proceedings of the IEEE/CVF Conference on Computer Vision and Pattern Recognition. pp. 10003--10015 (2024)

\bibitem{xie2026cari4d}
Xie, X., Wen, B., Chang, Y., Rabeti, H., Li, J., Yuan, Y., Pons-Moll, G., Birchfield, S.: Cari4d: Category agnostic 4d reconstruction of human-object interaction. In: Proceedings of the IEEE/CVF Conference on Computer Vision and Pattern Recognition. pp. 14006--14016 (2026)

\bibitem{xu2022vitpose}
Xu, Y., Zhang, J., Zhang, Q., Tao, D.: Vitpose: Simple vision transformer baselines for human pose estimation. Advances in neural information processing systems  \textbf{35},  38571--38584 (2022)

\bibitem{yang2024depth}
Yang, L., Kang, B., Huang, Z., Zhao, Z., Xu, X., Feng, J., Zhao, H.: Depth anything v2. Advances in Neural Information Processing Systems  \textbf{37},  21875--21911 (2024)

\bibitem{yang2026sam3dbody}
Yang, X., Kukreja, D., Pinkus, D., Fan, T., Park, J., Shin, S., Cao, J., Liu, J.W., Ugrinovic, N., Sagar, A., et~al.: Sam 3d body: Robust full-body human mesh recovery. In: Proceedings of the IEEE/CVF Conference on Computer Vision and Pattern Recognition. pp. 7209--7219 (2026)

\bibitem{yin2025progait}
Yin, X., Yang, B., Liu, W., Xue, Q., Alamri, A., Fiedler, G., Gao, W.: Progait: A multi-purpose video dataset and benchmark for transfemoral prosthesis users. In: Proceedings of the IEEE/CVF International Conference on Computer Vision. pp. 8984--8993 (2025)

\bibitem{ying2025ldpose}
Ying, J., Du, H., Zhang, K., Li, L., Yu, X.: Ldpose: Towards inclusive human pose estimation for limb-deficient individuals in the wild. In: Proceedings of the IEEE/CVF International Conference on Computer Vision. pp. 9865--9875 (2025)

\bibitem{zhang2025tapip3d}
Zhang, B., Ke, L., Harley, A., Fragkiadaki, K.: Tapip3d: Tracking any point in persistent 3d geometry. Advances in Neural Information Processing Systems  \textbf{38},  135284--135303 (2025)

\bibitem{zhang2026efficiently}
Zhang, C., Le~Moing, G., Koppula, S., Rocco, I., Momeni, L., Xie, J., Sun, S., Sukthankar, R., Barral, J.K., Hadsell, R., et~al.: Efficiently reconstructing dynamic scenes one d4rt at a time. In: Proceedings of the IEEE/CVF Conference on Computer Vision and Pattern Recognition. pp. 7382--7392 (2026)

\bibitem{zhang2025monst3r}
Zhang, J., Herrmann, C., Hur, J., Jampani, V., Darrell, T., Cole, F., Sun, D., Yang, M.H.: Monst3r: A simple approach for estimating geometry in the presence of motion. In: International Conference on Learning Representations. vol.~2025, pp. 82863--82886 (2025)

\bibitem{zheng20213d}
Zheng, C., Zhu, S., Mendieta, M., Yang, T., Chen, C., Ding, Z.: 3d human pose estimation with spatial and temporal transformers. In: Proceedings of the IEEE/CVF international conference on computer vision. pp. 11656--11665 (2021)

\bibitem{zhu2023motionbert}
Zhu, W., Ma, X., Liu, Z., Liu, L., Wu, W., Wang, Y.: Motionbert: A unified perspective on learning human motion representations. In: Proceedings of the IEEE/CVF international conference on computer vision. pp. 15085--15099 (2023)

\end{thebibliography}

\end{document}